\documentclass[preprint]{vgtc}                          

\graphicspath{{figures/}{pictures/}{images/}{./}} 

\usepackage{times}                     

\usepackage{tabu}                      
\usepackage{booktabs}                  
\usepackage{lipsum}                    
\usepackage{mwe}                       
\usepackage{enumitem}
\usepackage{mathptmx}                  

\onlineid{9891}

\vgtccategory{Research - Short Paper}

\vgtcinsertpkg

\newcommand{\ORNLCopyright}{%
  Notice: This manuscript has been authored by UT-Battelle, LLC, under contract DE-AC05-00OR22725 with the US Department of Energy (DOE). The US government retains and the publisher, by accepting the article for publication, acknowledges that the US government retains a nonexclusive, paid-up, irrevocable, worldwide license to publish or reproduce the published form of this manuscript, or allow others to do so, for US government purposes. DOE will provide public access to these results of federally sponsored research in accordance with the DOE Public Access Plan ( https://www.energy.gov/doe-public-access-plan ).%
}

\title{How Do Artificial Agents Make Decisions in Light of Visual Data?}

\author{Keyty Loza\thanks{e-mail: kloza1@lamar.edu }\\ %
        \scriptsize Lamar University %
\and Shayan Monadjemi\thanks{e-mail: monadjemis@ornl.gov\\ \\ \ORNLCopyright}\\ %
     \scriptsize Oak Ridge National Laboratory %
}

\teaser{
    \centering
    \includegraphics[clip, trim=0cm 3.5cm 0cm 3.65cm, width=0.78\textwidth]{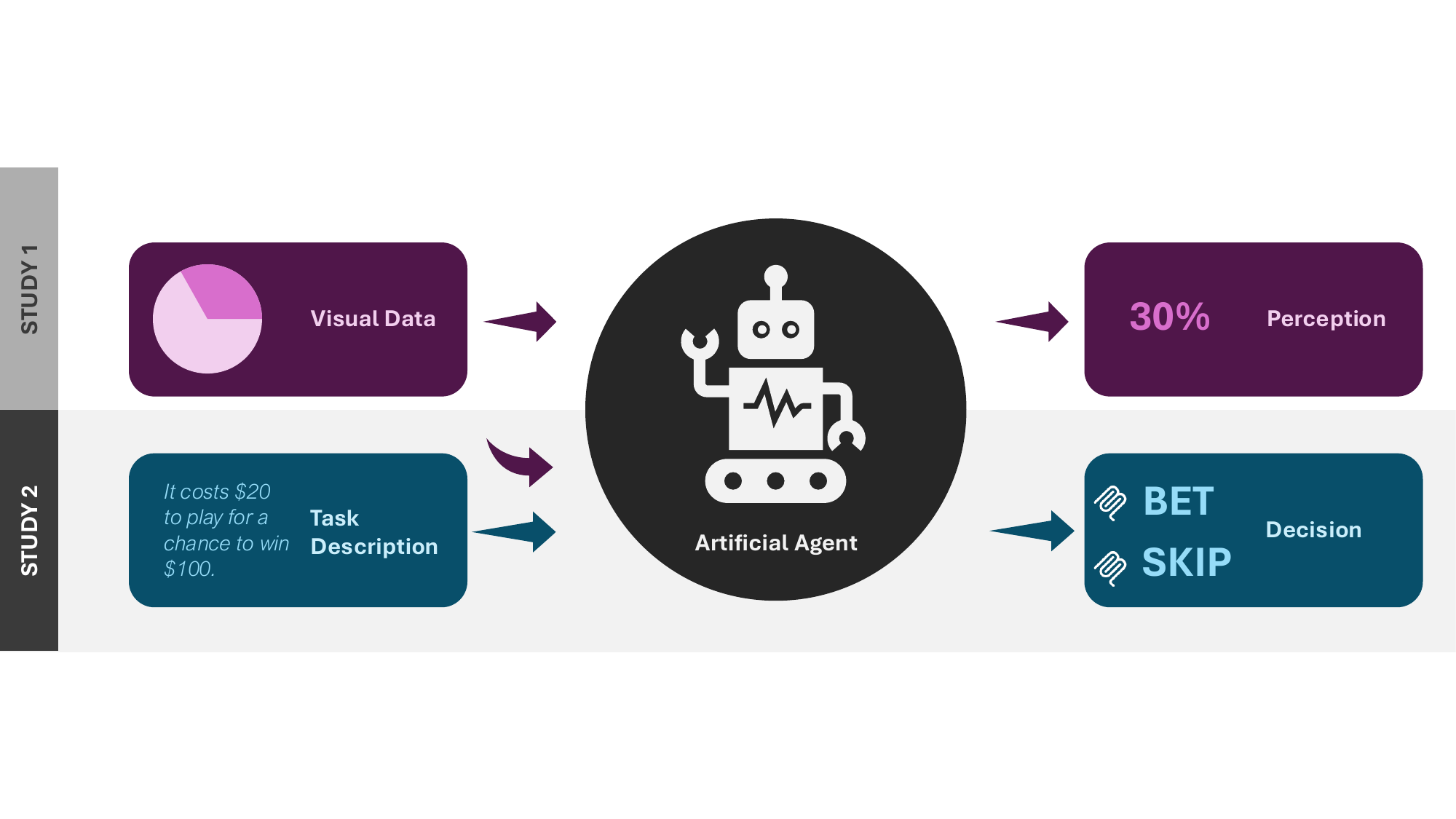}
    \caption{We investigate how artificial agents, powered by vision-language models, perceive uncertainty from data visualizations and how they make decisions (in the form of Model Context Protocol tool calls) based on the visual data.}
    \label{fig:teaser}
}

\abstract{
Vision-language models (VLMs) have demonstrated significant advances in reasoning with visual data, which is particularly important given that much of human knowledge is encoded visually in scientific articles and other sources. 
However, a gap remains in understanding how artificial agents, powered by VLMs, perceive and navigate decision-making under uncertainty in the presence of visual data. 
This study explores this gap by evaluating and analyzing autonomous decision-making processes.
Building upon prior work in visualization literacy and decision theory, we conducted two complementary studies: one examining how agents perceive and quantify uncertainty encoded in visualizations, and another investigating how artificial agents make decisions based on visual data in the form of images. 
We analyzed perception accuracy and decision performance across models and visualization types. 
Our results show that GPT-4o demonstrates the highest accuracy in perceiving uncertainty in pie and bar charts among the tested models. 
For decision-making, autonomous agents showed the best alignment with rational decision theory when uncertainty was represented through circle charts, icon arrays, or text.
} 


\nocopyrightspace

\begin{document}

\preprinttext{Presented at the Workshop on GenAI, Agents, and the Future of VIS,
IEEE VIS 2026.}


\firstsection{Introduction}

\maketitle
Artificial agents, powered by foundation models, are increasingly being delegated agency to make decisions based on heterogeneous information sources, including visual data. Despite demonstrated success in various domains, empirical evidence characterizing how artificial agents perceive visual data and make decisions guided by such data remains limited.
As a motivating example, consider emerging efforts to leverage AI for accelerating scientific discovery. 
Given that scientific knowledge is extensively captured as visual data in the literature, artificial agents' ability to accurately perceive and make informed decisions based on such visualizations is crucial.
Vision-Language Models (VLMs) have emerged as powerful tools for addressing this challenge. We conducted two studies to probe the ability of AI agents (powered by VLMs) to interpret uncertainty visualizations and make autonomous decisions. Building upon prior work in the VIS community on visualization literacy~\cite{lee2017vlat,pandey2023minivlat,wong-2018-towards} and decision theory~\cite{bancilhon2020gamble}, we treat artificial agents not as passive interpreters of visual data, but as active decision-makers.

In our first study, we present various levels of risk encoded in one of five visualization types (icon arrays, pie charts, circle charts, triangle charts, and bar charts) to four VLMs (Gemma-4, Granite3.2-Vision, GPT-4o, and Qwen3-VL:8b) and asked each model to report the perceived probability.
In our second study, we prompted an agent capable of initiating tool calls through the Model Context Protocol with a decision problem requiring a choice between betting or not betting based on visually presented uncertainty.
We tracked agent responses and decisions, deriving metrics such as perception error, expected reward, and decision alignment with rational choice theory. 
Our results indicate that artificial agents can accurately perceive risk from visualizations on average, with the most accurate perception emerging from pie and bar charts. Furthermore, we found that 80\% of decisions made by artificial agents based on visual data align with rational choice theory, whereas 13\% are risk-averse and 7\% are risk-seeking. These results represent an initial step in evaluating artificial agents as active decision-makers and understanding how data visualization design influences their decisions.

In summary, our key contributions are as follows:
\begin{itemize}[noitemsep]
    \item A visualization literacy study adapted from prior work to evaluate artificial agents' ability to estimate probabilistic information encoded in visual representations.
    \item An evaluation of four VLMs across five visualization types, demonstrating that most models achieve reasonable accuracy in probability estimation.
    \item An evaluation of autonomous decision-making by an AI agent based on images of visual data, showing that visualization type influences the likelihood of rational versus risk-seeking/risk-averse decisions.
\end{itemize}
     
\section{Background}

Data visualization is a prominent method of communicating risk and uncertainty to decision makers.
Prior work has primarily focused on how \emph{humans} perceive risk from visualizations and how their decisions are influenced by visual data.
With increasing delegation of agency to AI agents, researchers have begun investigating similar concepts for artificial agents, as opposed to humans. In this section, we give an overview of prior work on visualization literacy and decision theory, two areas that can help us unravel how artificial agents may initiate actions in light of visual data.

\subsection{Visualization Literacy}

Different populations and user groups consume data visualizations for various reasons~\cite{hosseini-2025-what, wong-2018-towards, ziemkiewicz-2012-understanding}. For example, scientists are presented with scientific findings in the form of visualizations, business analysts use visualizations to monitor revenue, and the general public uses visualizations while making medical decisions. It is, therefore, critical to assess people's abilities to accurately understand visual data and incorporate them into their decisions.
To this end, prior work has proposed a set of questions referred to as Visualization Literacy Assessment Test~\cite{lee2017vlat, pandey2023minivlat} to assess people's ability to understand visualizations.

With increasing adoption of agentic workflows, Dong and Crisan~\cite{dong-2026-probing} and Pandey et al.~\cite{pandey-2025-benchmarking} conducted evaluation studies where various vision-language models (VLMs) complete the visual literacy tests. Their results confirm VLMs' ability to comprehend data visualizations and answer questions about the charts. In this work, we build upon prior work by investigating how VLMs perceive uncertainty visualizations and make informed decisions, where the action set defining the decision space are provided to artificial agents in the form of Model Context Protocol (MCP) tools.

\subsection{Decision Theory}
Several factors, including stress, time constraints, and personal circumstances, influence human decision-making~\cite{bancilhon2025anatomy,lee2017vlat,pandey2023minivlat}. AI agents are not subject to these human-specific factors; however, they face different challenges, such as accurately perceiving uncertainty in data visualizations. 
While decision-making is a nuanced process, various theories help us characterize decisions. Throughout this work, we utilize the notion of \emph{rationality}, which states that a decision is rational when the expected benefits outweigh the costs.
In the lottery scenario used in this work, agents must decide whether to ``bet'' or ``do not bet'' based on the perceived probability, with no neutral option available. The agent analyzes the visualization, infers the probability of winning, reasons about the presented information, and reaches a conclusion before making its final decision. Consider the following prompt as an example:
\begin{quote}
    It costs 60 dollars to play the game, and you have an 80\% chance of winning 100 dollars. Given this information, make a decision.
\end{quote}
In this case, the expected return of the decision is:
\begin{equation}
    (0.8 \times \$100) - \$60
\end{equation}
Since this value is positive, the rational decision according to economic theory is to pay the price and play this game.

\section{Study Design}

\begin{figure*}[!ht]
    \centering
    \includegraphics[width=0.9\linewidth]{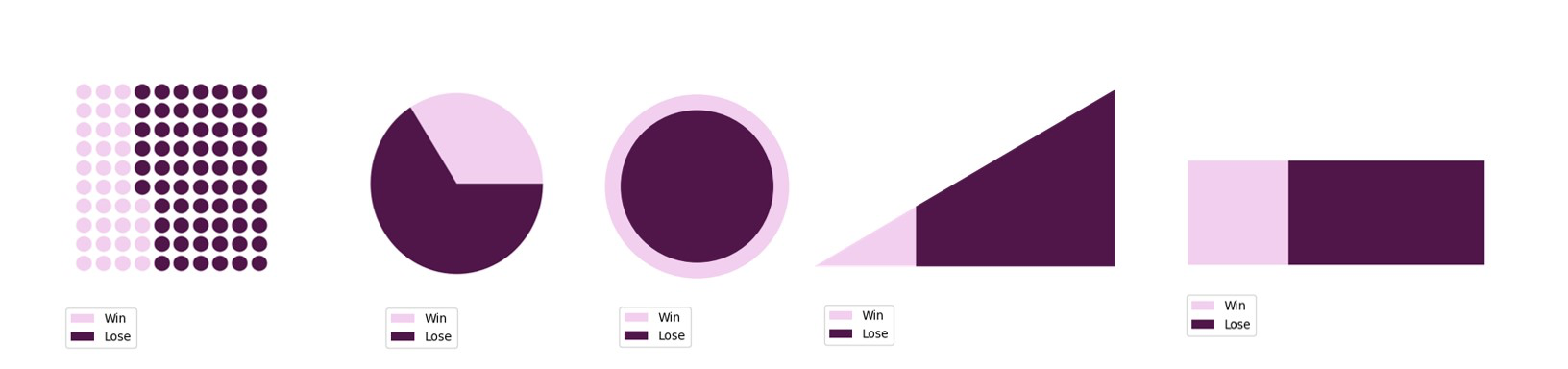}
    \caption{Visualization types considered in this study are icon array, pie charts, circle, triangle and bar charts.}
    \label{fig:vis-types}
\end{figure*}

This study extends previous work to investigate how AI agents make decisions under visual uncertainty. To this end, we generate lottery scenarios as a proxy for agentic decision scenarios with the following parameters: 
\begin{itemize}[noitemsep]
    \item \textbf{Chance of winning} -- from 5\% to 100\%, with 5\% increments.
    \item \textbf{Visualization type} -- the five options from Figure~\ref{fig:vis-types}.
    \item \textbf{Potential prize} -- kept constant at \$100.
    \item \textbf{Cost of playing} -- from \$10 to \$100, with \$10 increments.
\end{itemize}
We repeat each experiment 5 times to ensure the observations are consistent.

\subsection{Study 1: Probability Estimation}

In this study, our goal is to assess whether VLMs can accurately estimate a probability from visualizations. Hence, we limit the experiment space to \textbf{chance of winning} and \textbf{visualization type} only. We select four VLMs (Gemma-4, Granite3.2-Vision, GPT-4o, and Qwen3-VL:8b) and prompt each of them with the following:
\begin{quote}
    Based on the visualization, what is the probability of winning? You may only respond with a number between 0 to 1, with up to two significant digits. Do not include text, descriptions, or multiple numbers in your response.
\end{quote}

\subsection{Study 2: Decision}

Next, we conducted another study where the agents are responsible for making decisions based on the visualization image. As opposed to Study 1 where agents simply report their perceived probability of winning, here they had to initiate an action available to them through a simple Model Context Protocol (MCP) server. The available actions were:
\begin{itemize}[noitemsep]
    \item \textbf{Bet} -- choose to play the game.
    \item \textbf{Do not bet} -- choose not to play the game.
\end{itemize}

We narrowed our set of models to only GPT-4o in this study due to its ability to both make tool calls and process images, and due to its availability as a scalable service to us. We defined agents with the model, set of tools, and the following system prompt defining the agent's persona:
\begin{quote}
    You are an agent deciding whether to play a game or not. Given all the data and information about the game, use your tools to make a decision to bet or not bet.
\end{quote}
Then, we prompted the agent with a visualization and the following request template:
\begin{quote}
    It costs $c$ dollars to play the game, and you have the potential to win \$100 dollars. Given the chance of winning presented in the visualization, make a decision. You must select an action based on this information.
\end{quote}

We recorded the agent's decisions (i.e., tool calls) and the verbal response, if available.
We hypothesized that agents, powered by VLMs, will make rational decisions when presented with uncertaincy values encoded as visualizations.

\section{Results}

\subsection{Probability Estimation}

\begin{figure*}[!ht]
    \centering
    \includegraphics[width=0.4\linewidth]{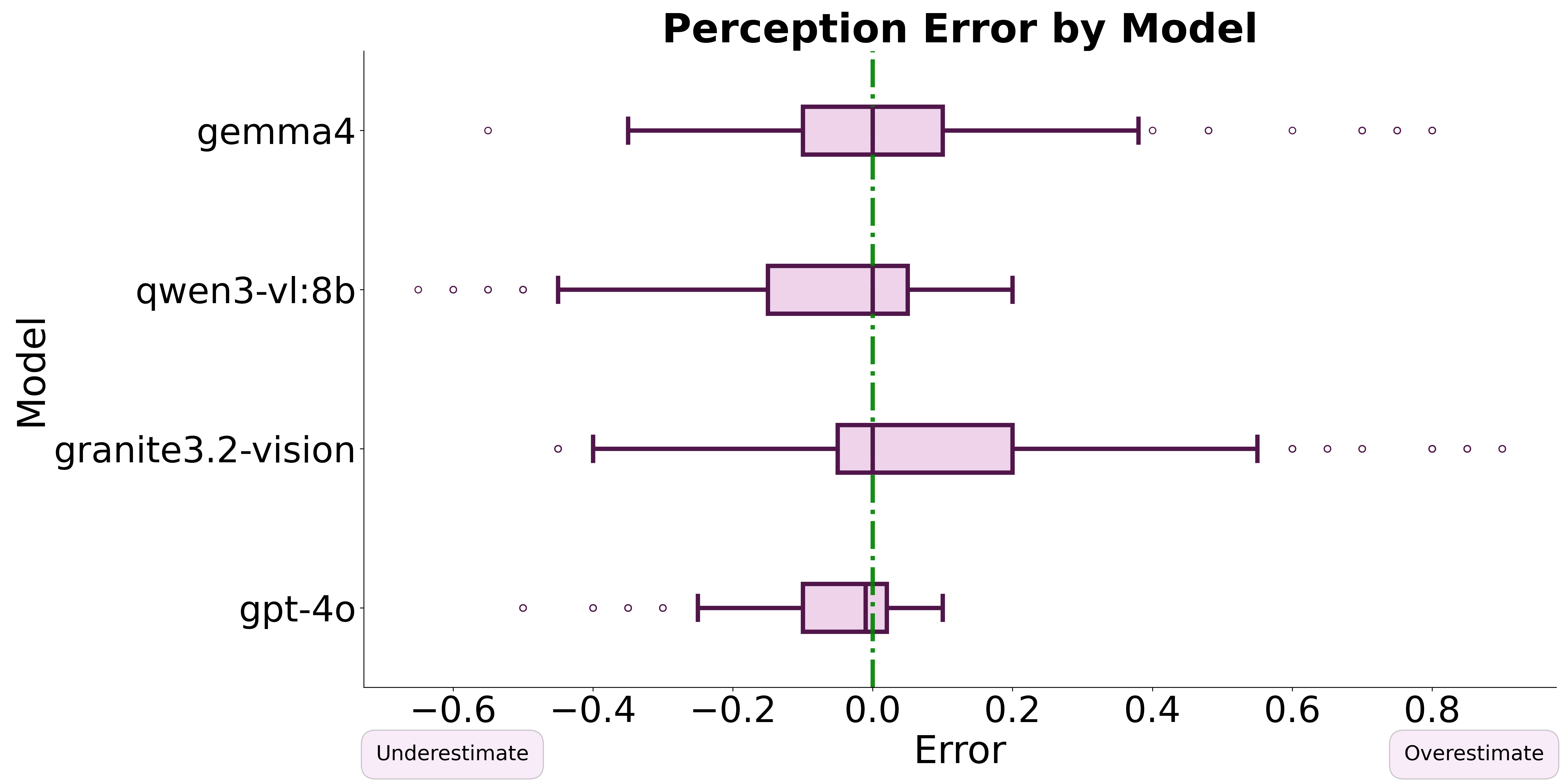}
    \hspace{0.08\linewidth}
    \includegraphics[width=0.4\linewidth]{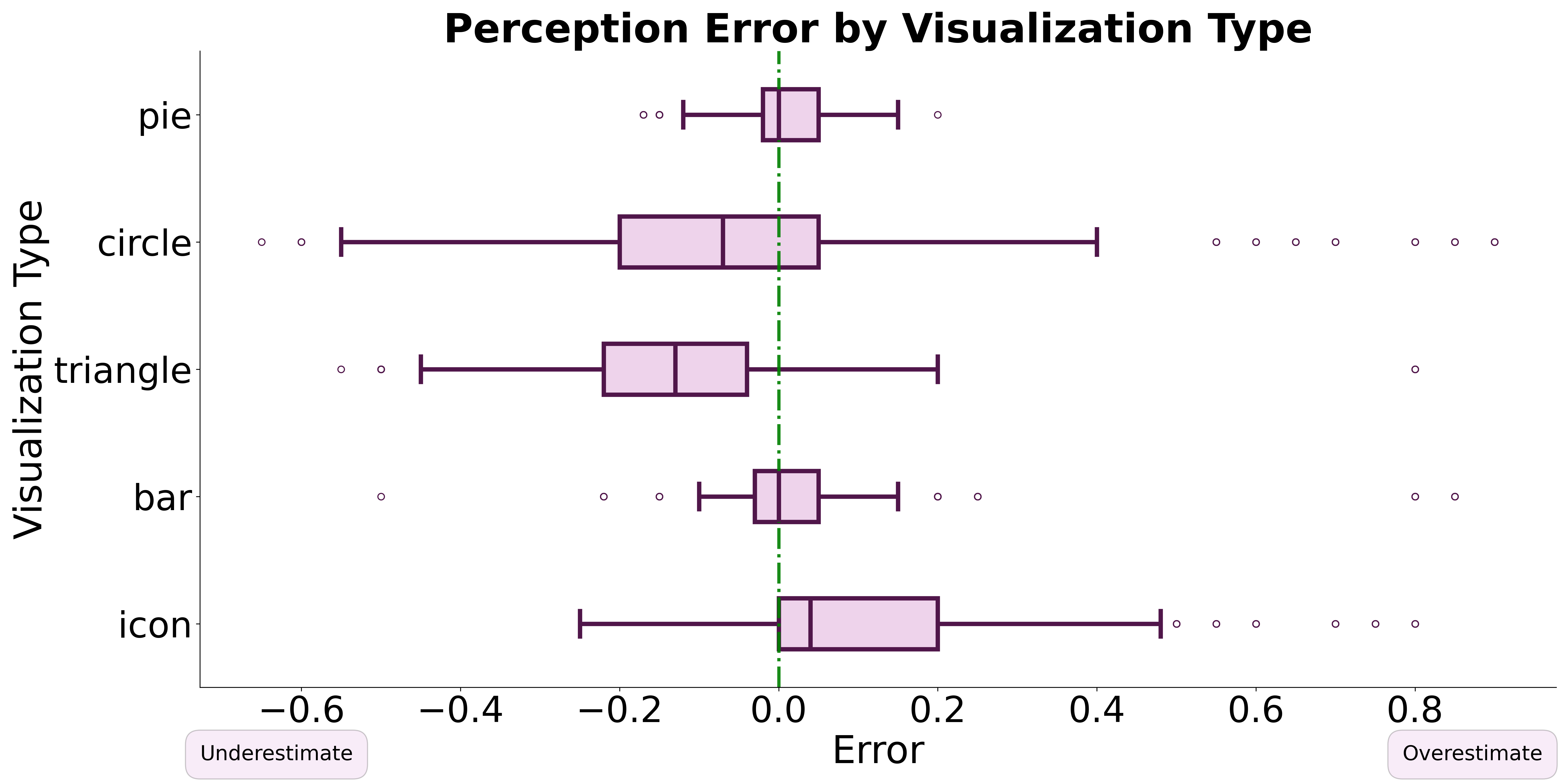}
    \caption{Left: The distribution of error per model. Right: The distribution of error per visualization type. Most charts were, on average, underestimated. Icon arrays were the only charts to be overestimated on average.}
    \label{fig:placeholder}
\end{figure*}

The evaluated models demonstrate the ability to perceive visual point-estimates, though performance varied across architectures. To quantify model behavior and analyze error perception, we examined the distribution of perceptual errors across the VLMs using box plots. Distributions centered closer to the zero-error baseline indicate greater perceptual accuracy (Figure 3), while distributions shifted away from zero indicate systematic underestimation or overestimation. Wider distributions indicate greater variability in each model's probability perception.
To obtain perception accuracy, we calculated the mean absolute error (MAE) using the following formula:
\begin{equation}
    MAE = |p_{perceived} - p_{actual}|
\end{equation}
For the mean perceived error across models and visualization types, we used the following formula:
\begin{equation}
    Error = p_{perceived} -  p_{actual}
\end{equation}

A comparison of model performance across visualization types reveals substantial differences in both probability estimation accuracy and consistency. GPT-4o achieved the highest accuracy with a MAE of 0.083, while Qwen3-VL:8b and Gemma-4 achieved MAEs of 0.118 and 0.128, respectively. Granite3.2-Vision had the highest error rate (0.203 MAE), indicating lower overall accuracy across the evaluated visualizations. These results indicate that GPT-4o maintained the most accurate performance. Analysis of average error reveals that Gemma-4 slightly overestimates probabilities (see Figure 3), while Granite3.2-Vision showed a stronger tendency toward overestimation ($\mu=0.092$). In contrast, GPT-4o ($\mu=-0.056$) and Qwen3-VL:8b ($\mu=-0.060$) generally underestimated probabilities. Across all tested models, overestimation and underestimation errors were balanced, suggesting no systematic bias in perceiving probabilities from visual data.

Performance on pie charts was particularly strong, consistent with previous findings~\cite{bancilhon2020gamble}, achieving a MAE near zero with low variability. Bar charts also demonstrated strong performance (0.078 MAE). Icon arrays, however, were generally interpreted incorrectly, with a tendency toward overestimation. In contrast, circle and triangle visualizations were consistently underestimated, negatively affecting error perception (Figure 3). Overall, these results demonstrate that the models can reason from visual representations of probability; however, their accuracy and consistency depend strongly on both visualization type and model architecture. This variability highlights the importance of considering visualization design when using VLMs for automated decision-making under uncertainty.

While no single visualization format achieved perfect zero-error perception across all models (Figure 3), most charts were underestimated, confirming that VLMs can reason but also make errors~\cite{dong2025probing}. Pie charts and bar charts performed closest to the baseline, while triangle charts showed the highest underestimation across all tested models. Icon arrays were the only visualization type consistently overestimated by models, suggesting they may be perceived as more risky due to contextual differences~\cite{bancilhon2020gamble}. This analysis provides critical insight: while no single visual format is perfectly interpreted by these agents, clear performance gaps persist across chart types.

\subsection{Decisions}
\begin{figure*}[!ht]
    \centering
    \includegraphics[width=0.9\linewidth]{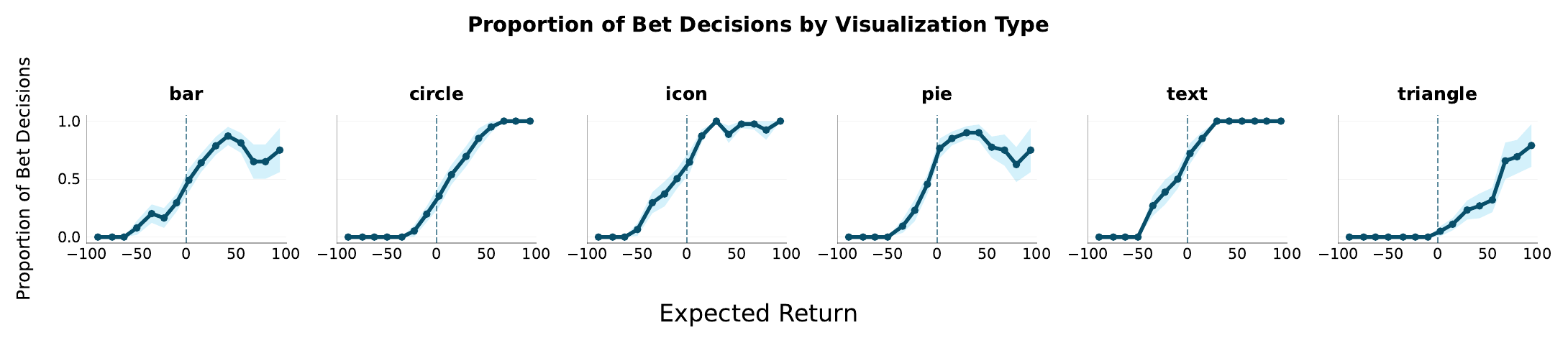}
    \caption{Proportion of rational decisions made across expected returns. Vertical line denotes the decision boundary for a rational agent.}
    \label{fig:decisions-1}
\end{figure*}

Our evaluation results illustrate how GPT-4o makes decisions under visual uncertainty. The model was tasked with identifying when it was rational to \textit{bet} and when it was rational to \textit{do not bet}. The model successfully recognized probabilities from visual data and reasoned about the relationship between cost and potential winnings in Figure 2. Text-based scenarios exhibit a relatively sharp transition around the rational decision boundary, whereas visual representations show varying response patterns. When making decisions, agents perceived the visualization and calculated an approximation of winning and losing outcomes. In lottery scenarios, humans typically show a preference for higher probability gains; agents exhibited similar behavior~\cite{bancilhon2020gamble}. Across visualization types, the proportion of \textit{BET} decisions generally increases with expected return, though this relationship varies by chart type. Circle charts, icon arrays, and text (no visualization) enabled more rational decisions compared to other visualization types. In Figure~\ref{fig:decisions-1}, we observed a drop in rational decisions for higher ends of expected returns for bar and pie charts, which was unexpected given their performance in Study 1.

\begin{figure}
    \centering
    \includegraphics[width=0.8\linewidth]{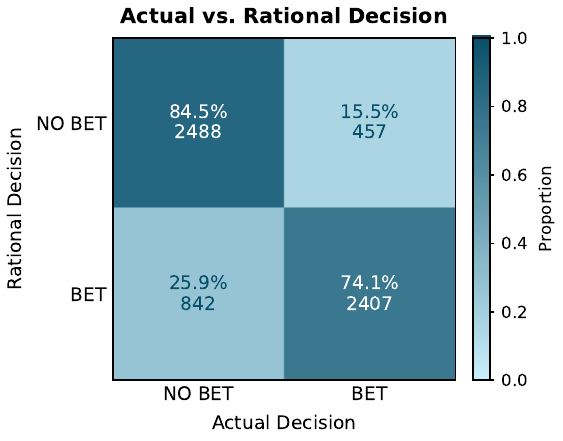}
    \caption{Rationality of decisions made by the artificial agent across experiments.}
    \label{fig:decisions-2}
\end{figure}

Figure~\ref{fig:decisions-2} compares the agents' actual decisions with the rational decisions defined by the experimental framework. When the rational decision was \textit{NO BET}, the agents reached the correct decision in 84.5\% of cases, while 15.5\% incorrectly chose \textit{BET}. When the rational decision was \textit{BET}, the agents reached the correct decision in 74.1\% of cases, while 25.9\% incorrectly chose \textit{NO BET}. These results indicate consistent decision performance, with agents more frequently making rational decisions in \textit{NO BET} scenarios than in \textit{BET} scenarios, demonstrating that VLMs are capable of making reasonable decisions under visual uncertainty.

\section{Discussion}
Consistent with the findings discussed above, our results demonstrate that artificial agents, powered by VLMs, can extract probabilistic information from data visualizations and use this information to make decisions under uncertainty. The direction and magnitude of probability estimation error provide insight into the impact of visualization design and model choice on the outcomes, with values closer to the zero-error baseline indicating greater accuracy. The differences observed across models suggest that visual reasoning capabilities are not uniform, even when agents are presented with the same underlying probability information. More importantly, these perceptual differences can extend to the agents' subsequent decisions. Although the same underlying data and decision criteria were maintained across visualization types, the agent exhibited different \textit{BET} and \textit{DO NOT BET} behaviors depending on how the information was visually represented. This suggests that visual encoding can influence not only the probability estimated by an agent but also how that information is used to reach a decision. This behavior is particularly relevant for autonomous decision-making systems, where errors introduced during visual interpretation may propagate through the reasoning process and ultimately affect the action selected by the agent.

The tendency of several models to underestimate displayed probabilities suggests that visual probability estimation remains imperfect in current VLMs. Conversely, the relatively strong performance observed for bar and pie charts indicates that certain visual encodings communicate probabilistic information to VLMs more effectively than others. These findings have important implications for the visualization community and may inform the development of VLM-friendly visual designs. The larger errors and variability observed for other visualization types, such as triangles and icon arrays, indicate greater difficulty in accurately interpreting the represented probabilities. Notably, circle visualizations exhibited different interpretations across models, suggesting that the same visual representation may not be consistently interpreted by different VLM architectures. These perceptual differences are particularly relevant to downstream decision-making, as inaccurate probability estimates can alter whether an agent selects \textit{BET} or \textit{DO NOT BET}. Therefore, evaluating VLMs solely on their ability to interpret visual data may not fully capture their reliability; both perceptual accuracy and the resulting decision behavior should be considered when evaluating VLM agents for autonomous decision-making under uncertainty.

\section{Limitation}
Our study is limited to a selected set of VLMs and visualization types, which may restrict the generalizability of the observed accuracy patterns. Although we are motivated by applications where artificial agents reason about complex scientific visualizations and make decisions in light of them, this study is a simplified first step towards understanding how artificial agents make decisions in light of visual uncertainty. For example, the experiments used controlled visualizations representing known probabilities, whereas real-world visualizations often contain greater complexity, including multiple variables, annotations, and contextual information. Although our evaluation identifies differences in probability estimation accuracy across chart types, it does not isolate the underlying mechanisms responsible for these differences. Evaluation of additional vision-language models would provide further insight into rational decision-making patterns. Future studies should expand the range of models, visualization designs, and experimental conditions to further examine the factors influencing autonomous decision-making.

\section{Future Work}

Future studies could expand beyond the current scope by investigating decision-making across a broader range of VLMs and visualization types, including distribution-based visualizations (e.g., violin plots, density plots). Such expansion would provide deeper insight into how different visual representations of uncertainty influence probability perception, reasoning, and agent decisions. Additionally, future research should evaluate how varying monetary stakes (gains and losses of different magnitudes) influence AI agent decision-making behavior in lottery scenarios.

\section{Conclusion}
This study highlights the importance of visualization design when integrating artificial agents, powered by VLMs, into autonomous decision-making systems. Our findings demonstrate that both model architecture and visualization type influence probability estimation accuracy and subsequent decision behavior.

\acknowledgments{
This project was supported in part by the 
U.S. Department of Energy, Office of Science, Office of Workforce Development for Teachers and Scientists (WDTS) under the Community College Internships Program (CCI).
This research used resources from the ORNL Research Cloud Infrastructure at the Oak Ridge National Laboratory, which is supported by the Office of Science of the U.S. Department of Energy under Contract No. DE-AC05-00OR22725. Research supported, in part, by the Laboratory Directed Research and Development Program of Oak Ridge National Laboratory, managed by UT-Battelle, LLC, for the US Department of Energy.
}

\bibliographystyle{abbrv-doi}

\bibliography{paper}
\end{document}